\documentclass[10pt,twocolumn,letterpaper]{article}

\usepackage{wacv}
\usepackage{times}
\usepackage{epsfig}
\usepackage{graphicx}
\usepackage{amsmath}
\usepackage{amssymb}
\usepackage{booktabs}
\usepackage{multirow}
\usepackage{xspace}
\usepackage{bm}
\usepackage{soul}

\usepackage[utf8]{inputenc}
\usepackage{enumitem}
\usepackage{overpic}

\usepackage{microtype}

\newcommand{\ourmethod}{OGMM\xspace}

\def\wacvPaperID{917} 

\wacvfinalcopy 

\ifwacvfinal
\usepackage[breaklinks=true,bookmarks=false]{hyperref}
\else
\usepackage[pagebackref=true,breaklinks=true,colorlinks,bookmarks=false]{hyperref}
\fi

\begin{document}

\title{Overlap-guided Gaussian Mixture Models for Point Cloud Registration}

\author{Guofeng Mei\\
University of Technology Sydney\\
Sydney, Australia\\
{\tt\small guofeng.mei@student.uts.edu.au}
\and
Fabio Poiesi\\
Fondazione Bruno Kessler\\
Trento, Italy\\
{\tt\small poiesi@fbk.eu}
\and
Cristiano Saltori\\
University of Trento\\
Trento, Italy\\
{\tt\small cristiano.saltori@unitn.it}
\and
Jian Zhang\\
University of Technology Sydney\\
Sydney, Australia\\
{\tt\small jian.zhang@uts.edu.au}
\and
Elisa Ricci\\
University of Trento $\&$ FBK\\
Trento, Italy\\
{\tt\small e.ricci@unitn.it}
\and
Nicu Sebe\\
University of Trento\\
Trento, Italy\\
{\tt\small niculae.sebe@unitn.it}
}

\maketitle
\thispagestyle{empty}

\begin{abstract}
Probabilistic 3D point cloud registration methods have shown competitive performance in overcoming noise, outliers, and density variations. 
However, registering point cloud pairs in the case of partial overlap is still a challenge. 
This paper proposes a novel overlap-guided probabilistic registration approach that computes the optimal transformation from matched Gaussian Mixture Model (GMM) parameters.
We reformulate the registration problem as the problem of aligning two Gaussian mixtures such that a statistical discrepancy measure between the two corresponding mixtures is minimized. 
We introduce a Transformer-based detection module to detect overlapping regions, and represent the input point clouds using GMMs by guiding their alignment through overlap scores computed by this detection module.
Experiments show that our method achieves superior registration accuracy and efficiency than state-of-the-art methods when handling point clouds with partial overlap and different densities on synthetic and real-world datasets. https://github.com/gfmei/ogmm
\end{abstract}

\vspace{-0.5cm}
\section{Introduction}\label{sec1}
With the rise of inexpensive 3D sensors for both indoor (e.g.~Microsoft Kinect) and outdoor (e.g.~LiDAR) scenes, the point cloud has become an important data source, which presents rich 3D spatial information efficiently. 
To produce large-scale point clouds, 3D point cloud registration has been widely investigated.
The point cloud registration refers to the problem of finding a rigid relative pose transformation that aligns a pair of point clouds into the same coordinate frame~\cite{huang2020feature,mei2021point}.
The registration quality can directly impact applications such as 
robotics~\cite{wang2020,Zhou2022}, 
augmenting reality~\cite{borrmann2018large}, 
autonomous driving~\cite{nagy2018real,poiesi2022}, 
and radiotherapy~\cite{li2019noninvasive,ma2018point}.
However, sensor noise, varying point densities, outliers, occlusions, and partial views are challenges that still affect the performance of these applications in the real world~\cite{kadam2022r}.

Point cloud registration approaches can be broadly categorized into correspondence-free and correspondence-based~\cite{mei2021point}. 
\textit{Correspondence-free registration approaches} aim at minimizing the difference between the global features extracted from two input point clouds~\cite{mei2021point,huang2020feature,aoki2019pointnetlk}.
These global features are typically computed based on all the points of a point cloud, making correspondence-free approaches inadequate to handle scenes with partial overlaps, such as those captured in the real world \cite{zhang2020deep,choy2020deep}. 
\textit{Correspondence-based registration approaches} rely on point-level correspondences between two input point clouds~\cite{fu2021robust,wang2019deep,besl1992method}.
Despite showing promising results, these approaches suffer from two major challenges: 
i) real-world point clouds do not contain exact point-level correspondences due to sensor noise and density variations \cite{huang2022unsupervised,yuan2020deepgmr,sun2022probability}; 
ii) the size of the correspondence search space increases quadratically with the number of points of the two point clouds \cite{yuan2020deepgmr}. 
An alternative to finding point-to-point correspondences is using distribution-to-distribution matching through probabilistic models~\cite{huang2022unsupervised,yuan2020deepgmr}.
These probabilistic registration techniques showed greater robustness to noise and density variations than their point-to-point counterpart \cite{sun2022probability}, however, they typically require their inputs to share the same distribution parameters (e.g., Gaussian Mixture Models). 
Due to this, they can only handle complete-to-complete~\cite{yuan2020deepgmr} or partial-to-complete~\cite{sun2022probability} point cloud registration setups. 
Partial-to-partial setups, which are typical in real-world applications, may have disjoint distribution parameters.
Therefore, when state-of-the-art approaches are used in these setups are highly likely to underperform.

In this paper, we propose an overlap-guided GMM-based registration method, named OGMM, to mitigate the limitations of partial-to-partial setups without using exact point-level correspondences.
We reformulate the problem of registering point cloud pairs as the alignment of two Gaussian mixtures by minimizing a statistical discrepancy measure between the two corresponding mixtures.
We introduce an overlap score to measure how likely points locate in the overlapping areas between source and target point clouds via a Transformer-based deep neural network.
The input point cloud is modeled through GMMs under the guidance of this overlap score.
The self-attention or cross-attention of Transformer networks come with computational and memory requirements that scale quadratically with the size of point clouds ($N^2$), hindering their applicability to large-scale point cloud datasets. 
Therefore, we introduce the idea of \textit{clustered attention}, which is a fast approximation of self-attention. 
Clustered attention groups a set of points into $J$ clusters and compute the attention for these clusters only, making the complexity linear with the number of clusters, i.e., $N\cdot J$, where $J << N$.
\ourmethod is inspired by DeepGMR~\cite{yuan2020deepgmr}, but it differs from it in two ways. 
First, our probabilistic paradigm can handle partial-to-partial point cloud registration problems through the overlap score constraint. 
Second, our network learns a consistent GMM representation across feature and geometric space rather than fitting a GMM in a single feature space.
We evaluate our approach on ModelNet40 \cite{wu20153d}, 7Scene \cite{shotton2013scene}, and ICL-NUIM \cite{handa2014benchmark}, comparing our approach against traditional and deep learning-based point cloud registration approaches.
We use the typical evaluation protocol for point cloud registration that is used in this literature \cite{fu2021robust}.
\ourmethod achieves state-of-the-art results and largely outperforms DeepGMR on all the benchmarks.

\vspace{.1cm}
\noindent In summary, the main contributions of this work are:
\setlist{nolistsep}
    \begin{itemize}[noitemsep]
    \item We propose a learning-based probabilistic registration framework under the guidance of overlap scores;
    \item We propose a cluster-based overlap Transformer module to embed cross point cloud information that enables the detection of overlap regions;
    \item We introduce a cluster-based loss to ensure that our network learns a consistent GMM representation across feature and geometric space rather than fitting a GMM in a single feature space;
    \item We achieve state-of-the-art accuracy and efficiency on a comprehensive set of experiments, including synthetic and real-world datasets.
\end{itemize}

\vspace{-.1cm}
\section{Related Work}\label{sec2}
\vspace{-.1cm}
We review both correspondence-free and correspondence-based point cloud registration methods, and Transformer-related works as it is a major component in our approach.
\vspace{-.2cm}
\subsection{Correspondence-free Registration}
\vspace{-.2cm}
The core idea of the correspondence-free registration approaches, such as PointNetLK \cite{aoki2019pointnetlk} and FMR \cite{huang2020feature}, is to first extract the global features from the source and target point clouds, and then regress the rigid motion parameters by minimizing the difference between global features of two input point clouds. 
Such methods do not require point-point correspondences and are robust to density variations.
These methods use extracted global features to reduce the point cloud's dimension so that the algorithm's time complexity does not increase as the number of points increases. 
However, they highly depend on fairly high overlaps (more than 90\%) between two point clouds and suffer from performance degradation in the case of partially-overlapping point clouds, which are typical in real-world scenarios~\cite{zhang2020deep,choy2020deep}.

\subsection{Correspondence-based Registration}

\subsubsection{Point-level Correspondence-Based Registration.}
The most popular point-to-point registration algorithm is ICP \cite{besl1992method}, which alternates between rigid motion estimation and the correspondences searching \cite{wang2019deep,huang2017coarse} by solving a $L_2 $-optimization. 
However, ICP converges to spurious local minima due to the non-convexity nature of the problem.
Based on this algorithm, many variants have been proposed. 
For example, the Levenberg-Marquardt ICP \cite{fitzgibbon2003robust} uses a Levenberg-Marquardt algorithm to yield the transformation by replacing the singular value decomposition with gradient descent and Gaussian-Newton approaches, accelerating the convergence while ensuring high accuracy. 
Go-ICP \cite{yang2015go} solves the point cloud alignment problem globally by using a Branch-and-Bound (BnB) optimization framework without prior information on correspondence or transformation. 
RANSAC-like algorithms are widely used for robust finding the correct correspondences for registration \cite{li2021point}. 
FGR \cite{zhou2016fast} optimizes a Geman-McClure cost-induced correspondence-based objective function in a graduated nonconvex strategy and achieves high performance. 
TEASER \cite{yang2020teaser} reformulates the registration problem as an intractable optimization and provides readily checkable conditions to verify the optimal solution. 
These methods still face challenges when point clouds are affected by noise, outlier, and density variations~\cite{zhang2020deep}.

Recently, several deep features \cite{huang2020feature,zhang2020deep} are proposed to accurately estimate the point-level correspondences. 
For instance, DCP \cite{wang2019deep} employs DGCNN \cite{wang2019dynamic} for feature extraction and an attention module to generate soft matching pairs. 
RPMNet \cite{yew2020rpm}  proposes a method to solve the point cloud partial visibility by integrating the Sinkhorn algorithm into a network to get soft correspondences from local features. 
Soft correspondences can increase the robustness of registration accuracy.
RGM \cite{fu2021robust} utilizes a Transformer to aggregate information by generating soft graph edges for point-wise matching. 
IDAM \cite{li2019iterative} incorporates both geometric and distance features into the iterative matching process. RIENet \cite{shen2022reliable} propose a method to identify the inlier based on the graph-structure difference between the neighborhoods. 
Although achieving remarkable performance, most of these approaches rely on point-to-point correspondences, and thus they are still sensitive to noise and density variations~\cite{huang2022unsupervised}.

\subsubsection{Probabilistic Registration.}
The probabilistic registration methods model the distribution of point clouds as a density function often via the use of GMMs and perform alignment either by employing a correlation-based approach or using an EM-based optimization framework~\cite{yuan2020deepgmr,lawin2018density}. 
A commonly used formulation, such as CPD \cite{myronenko2010point} and FilterReg \cite{gao2019filterreg}, represents the geometry of the target point cloud using a GMM distribution over 3D Euclidean space. 
The source point cloud is then fitted to the GMM distribution under the maximum likelihood estimation (MLE) framework. 
Another statistical approach, including GMMReg \cite{jian2010robust}, JRMPC \cite{evangelidis2017joint}, and DeepGMR \cite{yuan2020deepgmr}, builds GMM probability distribution on both the source and the target point clouds. 
These methods show robustness to outliers, noise, and density variation~\cite{yuan2020deepgmr}. These models usually assume that both source and target point clouds share the same GMM parameters or that one of the point clouds is supposed to be ``perfect", thus leading to biased solutions as long as both point clouds contain noise and outliers (such as partial-to-partial registration) \cite{evangelidis2017joint}.
Considering the above factors, we propose an overlap-guided 3D point cloud registration algorithm based on Gaussian Mixture Model (GMM) that can be applied to partial-to-partial registration problems.

\subsection{Transformers in 3D Point Clouds}
Due to the inherent permutation invariance and strong global feature learning ability, 3D Transformers are well suited for point cloud processing and analysis. 
They have surpassed state-of-the-art non-Transformer algorithm performance. 
In particular, A-SCN~\cite{xie2018attentional} is the first such example of a Transformer-based approach for point cloud learning. Later, PCT~\cite{guo2021pct}, which is a pure global Transformer network, generates positional embedding using 3D coordinates of points and adopts a transformer with an offset attention module to enrich features of points from its local neighborhood. PointTransformer~\cite{zhao2021point} to construct self-attention networks for general 3D tasks with nearest neighbor search. However, they suffer from the fact that as the size of the feature map increases, the computing and memory overheads of the original Transformer increase quadratically. Efforts to reduce the quadratic complexity of attention mainly focused on self-attention. For instance, PatchFormer~\cite{zhang2022patchformer} reduces the size of the attention map by first splitting the raw point cloud into patches, and then aggregating the local feature in each patch to generate an attention matrix. FastPointTransformer proposes centroid-aware voxelization and devoxelization techniques to reduce space complexity.
However, these works are less suitable for feature matching, where we are required to perform self- and cross-attention on features within and between point clouds, respectively. To this end, we propose clustered attention which groups a set of points into $J$ clusters and computes the attention only for these clusters, resulting in linear complexity for a fixed number of clusters.

\vspace{-0.2cm}
\section{Our approach}

\begin{figure*}[t]
\vspace{-.4cm}
\centering
\includegraphics[width=0.98\textwidth]{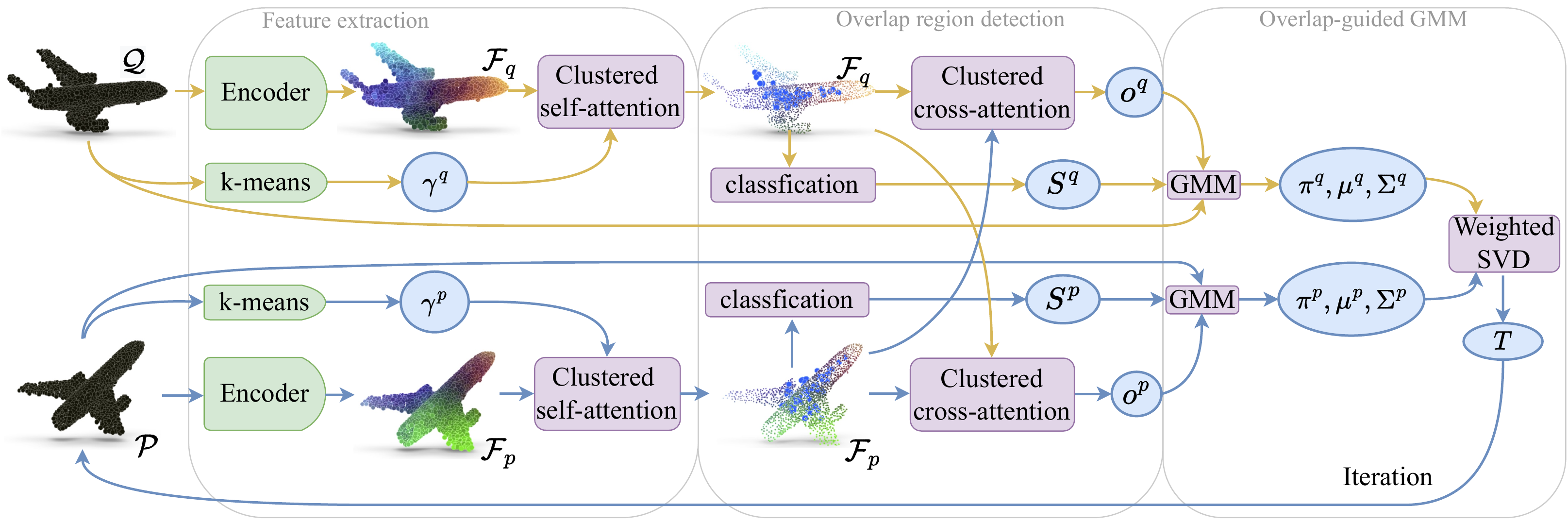}
\vspace{-.1cm}
\caption{
\ourmethod consists of three modules: feature extraction, overlap region detection, and overlap-guided GMM for registration. 
The shared weighted encoder extracts point-level features $\bm{\mathcal{F}}_p$ and $\bm{\mathcal{F}}_q$ from point clouds $\bm{\mathcal{P}}$ and $\bm{\mathcal{Q}}$, respectively.
The self-attention module updates the point-wise features $\bm{\mathcal{F}}_p$ and $\bm{\mathcal{F}}_q$. 
The overlap region detection module projects the updated features $\bm{\mathcal{P}}$ and $\bm{\mathcal{Q}}$ to overlap scores $\bm{o}_p, \bm{o}_q$, respectively.
$\bm{\mathcal{F}}_p$, $\bm{\mathcal{F}}_q$, $\bm{o}_p$ and $\bm{o}_q$ are used to estimate GMMs of $\bm{\mathcal{P}}$ and $\bm{\mathcal{Q}}$.
The weighted SVD is used to estimate the rigid transformation $T$ based on the estimated distributions.}
\label{fig:gmm}
\vspace{-.1cm}
\end{figure*}

Rigid point cloud registration aims to recover a rigid transformation matrix $T\in SE(3)$, which consists of rotation $R\in SO(3)$ and translation $\bm{t}\in \mathbb{R}^3,$ that optimally aligns the source point cloud $\bm{\mathcal{P}}=\{\bm{p}_i \in\mathbb{R}^{3}\big|i = 1, 2, ..., N\}$ to the target point cloud $\bm{\mathcal{Q}}=\{\bm{q}_j \in\mathbb{R}^{3}\big|j = 1, 2, ..., M\}$, where $N$ and $M$ represent the number of points in $\bm{\mathcal{P}}$ and $\bm{\mathcal{Q}}$, respectively.
Fig.~\ref{fig:gmm} illustrates our framework that consists of three modules: feature extraction, overlap region detection, and overlap-guided GMM for registration. 
The shared weighted encoder first extracts point-wise features $\bm{\mathcal{F}}_p$ and $\bm{\mathcal{F}}_q$ from point clouds $\bm{\mathcal{P}}$ and $\bm{\mathcal{Q}}$, respectively.
The clustered self-attention module then updates the point-wise features $\bm{\mathcal{F}}_p$ and $\bm{\mathcal{F}}_q$ to capture global context. 
Next, the overlap region detection module projects the updated features $\bm{\mathcal{P}}$ and $\bm{\mathcal{Q}}$ to overlap scores $\bm{o}_p, \bm{o}_q$, respectively.  
$\bm{\mathcal{F}}_p$, $\bm{\mathcal{F}}_q$, $\bm{o}_p$ and $\bm{o}_q$ are then used to estimate the distributions (GMMs) of $\bm{\mathcal{P}}$ and $\bm{\mathcal{Q}}$. Finally, weighted SVD is adopted to estimate the rigid transformation $T$ based on the estimated distributions.

\subsection{Feature Extraction}
The feature extraction network consists of a Dynamic Graph Convolutional Neural Network (DGCNN), positional encoding, and a clustered self-attention network. 
Given a point cloud pair $\bm{\mathcal{P}}$ and $\bm{\mathcal{Q}}$, DGCNN extracts their associated features $\bm{\mathcal{F}}_{p}=\{\bm{f}_{p_i} \in\mathbb{R}^{d}\big|i = 1, 2, ..., N\}$ and $\bm{\mathcal{F}}_{q}=\{\bm{f}_{q_j} \in\mathbb{R}^{d}\big|j = 1, 2, ..., M\}$. Here, $d=512$.

\vspace{-0.2cm}
\subsubsection{Attention module}
Transformer training and inference in previous works can be computationally expensive due to the quadratic complexity of self-attention over a long sequence of representations, especially for high-resolution correspondences prediction tasks. 
To mitigate this limitation, our novel cluster-based Transformer architecture operates after local feature extraction:
$\bm{\mathcal{F}}_p$ and $\bm{\mathcal{F}}_q$ are passed through the attention module to extract context-dependent point-wise features. 
Intuitively, the self-attention module transforms the DGCNN features into more informative representations to facilitate matching.

\vspace{.1cm}
\noindent\textbf{Spherical positional encoding.}
Transformers are typically fed with only high-level features, which may not explicitly encode the geometric structure of the point cloud~\cite{wang2019deep,huang2021predator}.
This makes the learned features geometrically less discriminative, causing severe matching ambiguity and numerous outlier matches, especially in low-overlap cases~\cite{qin2022geometric}.
A straightforward recipe is to explicitly inject positional encoding of 3D point coordinates, which assigns intrinsic geometric properties to the per-point feature by adding unique positional information that enhances distinctions among point features in indistinctive regions \cite{zhao2021point}. 
However, the resulting coordinate-based attentions are naturally transformation-variant \cite{qin2022geometric}, while registration requires transformation invariance since the input point clouds can be in arbitrary poses. 
To this end, we design spherical positional encoding, which leverages the distances and angles computed with the points, to encode the transformation-invariant geometric information of the points. 
Specifically, given a point $\bm{p}_i \in \bm{\mathcal{P}}$, we select the $k>0$ nearest neighbors $\mathcal{K}_i$ of $\bm{p}_i$ and compute the centroid $\bm{p}_c=\sum_{i=1}^{N}\bm{p}_i$ of mass of $\bm{\mathcal{P}}$. 
For each $\bm{p}_x\in \mathcal{K}_i$, we denote the angle between the vectors $\bm{p}_i-\bm{p}_c $ and $ \bm{p}_x-\bm{p}_c$ as $\alpha_{ix}$. 
The positional encoding $\bm{f}^{pos}_{p_i}\in\mathbb{R}^d$ of $\bm{p}_i$ is as
\begin{equation}\label{eq:pos}
	\bm{f}^{pos}_{p_i} = \varphi\left(\|\bm{p}_i-\bm{p}_c\|_2\right) +\max_{x\in \mathcal{K}_i}\{\phi\left(\alpha_{ix}\right)\},
\end{equation}
where $\varphi$ and $\phi$ are two MLPs, and each MLP consists of a linear layer and one ReLU nonlinearity function~\cite{xu2015empirical}.
Our positional encoding is invariant to rigid transformation since the distance and angles are invariant to transformation. We then updates features $\bm{\mathcal{F}}_{\bm{p}}$ of $\bm{\mathcal{P}}$ by $\bm{\mathcal{F}}_{\bm{p}}=\{\bm{f}^{pos}_{\bm{p}_i}+\bm{f}_{\bm{p}_i}\}$. The same operation is also applied in $\bm{\mathcal{Q}}$.

\vspace{.1cm}
\noindent \textbf{Cluster-based self-attention.}
The extracted local features have a limited receptive field which may hinder the distinctiveness of regions. 
Instead, humans can find correspondences between these regions not only based on the local neighborhood structures but also by exploiting context.
Self-attention is thus introduced to model global structures by establishing long-range dependencies.
We exploit this idea to improve the computational complexity of self-attention.
Specifically, we first use Wasserstein K-Means~\cite{fukunaga2021wasserstein} to cluster point cloud $\bm{\mathcal{P}}$ (or $\bm{\mathcal{Q}}$) into $J$ non-overlapping clusters in geometric space by $\bm{\gamma}^p\in \{0, 1\}^{N\times J}$ such that, $\gamma^p_{ij}=1$ (or $\gamma^q_{ij}=1$), if the $i$-th point of $\bm{\mathcal{P}}$ (or $\bm{\mathcal{Q}}$) belongs to the $j$-th cluster (denoted as $\bar{\bm{p}}_j$ or $\bar{\bm{q}}_j$) and 0 otherwise. Using this partitioning, we can now compute the clustered attention. 
First, we calculate the cluster centroids $\bm{f}_{\bar{p}_j}$ and $\bm{f}_{\bar{q}_j}$ of the points in each of these $J$ clusters in feature space as follows,
\vspace{-.2cm}
\begin{equation}\label{eq:fcentroids}
    \bm{f}_{\bar{p}_j} = \sum_{i=1}^N\frac{\gamma^p_{ij}\bm{f}_{p_i}}{\sum_k^N\gamma^p_{kj}}, \quad
    \bm{f}_{\bar{q}_j} = \sum_{i=1}^M\frac{\gamma^q_{ij}\bm{f}_{q_j}}{\sum_k^M\gamma^q_{kj}}.
\end{equation}
We then use a multi-attention layer with four parallel attention heads~\cite{wang2019deep} to update $\bm{\mathcal{F}}_{\bm{p}}$ in parallel via
\vspace{-.2cm}
\begin{equation}
    \bm{f}_{\bm{p}_i} \leftarrow \bm{f}_{\bm{p}_i} + \mbox{MLP}\left(\sum_{j=1}^J\alpha^{\bm{p}}_{ij}W^s_V\bm{f}_{\bar{\bm{p}}_j}\right),
    \vspace{-.2cm}
\end{equation}
where $\alpha^{\bm{p}}_{ij}$ is the element of matrix $\bm{\alpha}^{\bm{p}}=\mbox{Softmax}(\bm{S})$ with $\bm{S}=(W^s_Q\bm{f}_{\bm{p}_i})^\top W^s_K\bm{f}_{\bar{\bm{p}}_j}$. Here, $W^s_Q\in \mathbb{R}^{N\times d}, W^s_K\in \mathbb{R}^{J\times d}$ and $W^s_V\in \mathbb{R}^{J\times d}$ are the query, key and value matrices. The self-attention features for $\bm{\mathcal{Q}}$ are updated in the same
way. $\mbox{MLP}(\cdot)$ denotes a three-layer fully connected network with instance normalization \cite{ulyanov2016instance} and ReLU~\cite{xu2015empirical} activations after the first two layers.

\vspace{-0.2cm}
\subsubsection{Overlap Region Detection}

\noindent\textbf{Cluster-based cross-attention.}
Cross-attention is a typical module for point cloud registration tasks, which performs feature exchange between two input point clouds. 
Given the self-attention feature matrices $\bm{\mathcal{F}}_{\bm{p}}$ and $\bm{\mathcal{F}}_{\bm{q}}$ for $\bm{\mathcal{P}}$ and $\bm{\mathcal{Q}}$ respectively, we first update the cluster centroids $\bm{f}_{\bar{p}_j}$ and $\bm{f}_{\bar{q}_j}$ based on Eq.~\eqref{eq:fcentroids}. We denote the transformed features as $\bm{\mathcal{F}}^t_{p}$ and $\bm{\mathcal{F}}^t_{p}$ attained by cross-attention via
\vspace{-.2cm}
\begin{equation}
    \bm{f}^t_{\bm{p}_i} \leftarrow \bm{f}_{\bm{p}_i} + \mbox{MLP}\left(\sum\beta^{\bm{p}}_{ij}W^c_V\bm{f}_{\bar{\bm{q}}_j}\right),
\end{equation}
where $\beta^{\bm{p}}_{ij}$ is the element of matrix $\bm{\beta}^{\bm{p}}=\mbox{SoftMax}(\bm{C})$ with $\bm{C}=(W^c_Q\bm{f}_{\bm{p}_i})^\top W^c_K\bm{f}_{\bar{\bm{q}}_j}$. Here, $W^c_Q\in \mathbb{R}^{N\times d}, W^c_K\in \mathbb{R}^{J\times d}$ and $W^c_V\in \mathbb{R}^{J\times d}$ are the query, key and value matrices.  $\mbox{MLP}(\cdot)$ denotes a three-layer fully connected network with instance normalization \cite{ulyanov2016instance} and ReLU~\cite{xu2015empirical} activations after the first two layers. The same cross-attention block is also applied in reverse direction, so that information flows in both directions, $\bm{\mathcal{P}}\rightarrow \bm{\mathcal{Q}}$ and $\bm{\mathcal{Q}}\rightarrow \bm{\mathcal{P}}$.

\vspace{.1cm}
\noindent\textbf{Overlap score prediction.}
To deal with non-overlapping points, an overlap score prediction block is proposed. 
After obtaining the conditioned features $\bm{\mathcal{F}}^t_p$ and $\bm{\mathcal{F}}^t_q$, the per-point overlap score $\bm{o}_{{p}_i} \in [0, 1]$ can be computed by
\vspace{-.1cm}
\begin{equation*}
w_{ij} = \sigma\left({\bm{f}^t_{p_i}}^\top\bm{f}^t_{q_j}\big/{\tau}\right), 
\bm{o}_{p_i} = g_\beta\left(\mbox{cat}\left[\bm{f}^t_{q_i}, \bm{w}^{\top}_i g_\alpha \left(\bm{\mathcal{F}}^t_{q}\right)\right]\right),
\end{equation*}
where $\sigma$ is a \text{Softmax} function, and $\tau>0$ is a learned parameter that controls the soft assignment. When $\tau\rightarrow 0$, $w_{ij}$ converges to a hard nearest-neighbor assignment. $g_\alpha\left(\cdot\right): \mathbb{R}^d\rightarrow [0, 1]$ and $g_\beta\left(\cdot\right): \mathbb{R}^{d+1}\rightarrow [0, 1]$ are linear layers followed by an instance normalization layer and a sigmoid activation with different parameters $\alpha$ and $\beta$, respectively.

\subsection{Overlap-guided GMM for Registration}
A Gaussian mixture model (GMM) establishes a multimodal generative probability distribution over 3D space as a weighted sum of $L$ Gaussian densities~\cite{yuan2020deepgmr}, with the form
\vspace{-.2cm}
\begin{equation}\label{eq:density}
p(\bm{x})=\sum^L_{j=1}\pi_j\mathcal{N}\left(\bm{x}|\bm{\mu}_j,\bm{\Sigma}_j\right),\bm{x}\in \mathbb{R}^3.
\vspace{-.2cm}
\end{equation}
Each density $\mathcal{N}\left(\bm{x}|\bm{\mu}_j,\bm{\Sigma}_j\right)$ is referred to as a component of the GMM and parameterized by a mean $\bm{\mu}_j$ and covariance $\bm{\Sigma}_j$. The components are combined through a set of normalized mixing coefficients $\{\pi_1,\pi_2,\cdots,\pi_L\}$ that represent the prior probability of selecting the $j$-th component.

\vspace{.1cm}
\noindent\textbf{Learning posterior.}
We first apply a classification head $\phi_\theta$ that takes as input $\bm{\mathcal{F}}_p$ and $\bm{\mathcal{F}}_q$ and outputs joint log probabilities, and a softmax operator that acts on log probabilities to generate a probability matrix $\bm{S}^p$ and $\bm{S}^q$, respectively. 
The GMM parameters $\bm{\Theta}_p$ for point cloud $\bm{\mathcal{P}}$ consists of $L$ triples $(\pi^p_{j},\bm{\mu}^p_{j},\bm{\Sigma}^p_{j})$, where $\pi^p_{j}$ is a scalar mixture weight, $\bm{\mu}^p_{j}$ is a $3\times 1$ mean vector and $\bm{\Sigma}^p_{j}$ is a $3\times 3$ covariance matrix of the $j$-th component. 
For partial overlapping registration tasks, given the outputs $\bm{S}^p$ of $\phi_\theta$ together with the point coordinates $\bm{\mathcal{P}}$, and overlap scores $\bm{o}_p$, the GMM parameters can be written as
\vspace{-.2cm}
\begin{equation*}\label{eq:gmm}
    \begin{aligned}
    n_p &= \sum_{i=1}^N\bm{o}_{p_i}, \pi^p_{j} = \sum_{i=1}^N\frac{\bm{o}_{p_i}\bm{s}^p_{ij}}{\epsilon+n_p},
    \bm{\mu}^p_{j} = \sum_{i=1}^N\frac{\bm{o}_{p_i}\bm{s}^p_{ij}\bm{p}_i}{\epsilon+n_p\pi^p_{j}}, \\
    \bm{\Sigma}^p_{j} &= \frac{1}{\epsilon+n_p\pi^p_{j}}\sum_{i=1}^N\bm{o}_{p_i}\bm{s}^p_{ij}\left(\bm{p}_i-\bm{\mu}^p_{j}\right)\left(\bm{p}_i-\bm{\mu}^p_{j}\right)^\top,
    \end{aligned}
    \vspace{-.2cm}
\end{equation*}
where $\epsilon=1e-4$ is used to avoid zero in the denominator. By the same way, we can get the $\pi^q_{j}, \bm{\mu}^q_{j}$ and $\bm{\Sigma}^q_{j}$ for target point cloud, when giving $\bm{S}^q,\bm{o}_q$ and $\bm{\mathcal{Q}}$. The GMMs of point set $\bm{\mathcal{P}}$ and $\bm{\mathcal{Q}}$ are then given as
\begin{equation} \label{eq:post}
\begin{aligned}
    \mathbf{G}_{\mathcal{P}}\left(\bm{x}\right) &= \sum_{i=1}^L \pi^p_{j}\mathcal{N} \left(\bm{x}|\bm{\mu}^p_{j}, \bm{\Sigma}^p_{j}\right), \\
    \mathbf {G}_{\mathcal{Q}}\left(\bm{x}\right) &= \sum_{i=1}^L \pi^q_{j}\mathcal{N} \left(\bm{x}|\bm{\mu}^q_{j}, \bm{\Sigma}^q_{j}\right).
\end{aligned}
\end{equation}
We replace $\bm{p}_i$ and $\bm{q}_i$ with $\bm{f}_{p_i}$ and $\bm{f}_{q_i}$ to calculate the feature centroids $\{\bm{\nu}^p_{j}\}_{j=1}^L$ and $\{\bm{\nu}^q_{j}\}_{j=1}^L$ of $\bm{\mathcal{P}}$ and $\bm{\mathcal{Q}}$, respectively.

\vspace{.1cm}
\noindent\textbf{Estimating the transformation.}
Given the estimated GMMs parameters estimated through Eq.~\eqref{eq:post} as well as feature centroids $\{\bm{\nu}^p_{j}\}_{j=1}^L$ and $\{\bm{\nu}^q_{j}\}_{j=1}^L$, we first calculate the cluster-level matching matrix 
$\Gamma$ by solving the following optimal transport (OT) problem~\cite{peyre2019computational} as
\begin{equation}\label{eq:feature}
\begin{aligned}
& \min_{\Gamma}\sum_{i=1}^L\sum_{j=1}^L\Gamma_{ij}\|\bm{\nu}^p_{i}-\bm{\nu}^q_{j}\|_2^2,\\
& \mbox{s.t.,}~ \Gamma\bm{1}_M  = \bm{\pi}^{p}, \Gamma^\top\bm{1}_N = \bm{\pi}^{q}, \Gamma_{ij}\in[0, 1],
\end{aligned}
\end{equation}
where $\bm{\pi}^{t}=(\pi^{t}_1, \pi^{t}_1, \cdots, \pi^{t}_L), t\in\{p, q\}$. The minimization of Eq.~(\ref{eq:feature}) can be solved in polynomial time as a linear program, and we address this issue by adopting an efficient version of the Sinkhorn-Knopp algorithm~\cite{cuturi2013sinkhorn}. After obtaining the $\Gamma$, we then calculate the transformation by 
\begin{equation}
\min_{T}L\sum_{i=1}^L\sum_{j=1}^L\Gamma_{ij}\|T(\bm{\mu}^p_{i})-\bm{\mu}^q_{j}\|.
\end{equation}
Finally, we can solve $T$ in a closed-form using a weighted version of the SVD solution~\cite{yuan2020deepgmr}.

\begin{figure}[t]
    \centering
    \includegraphics[width=\columnwidth]{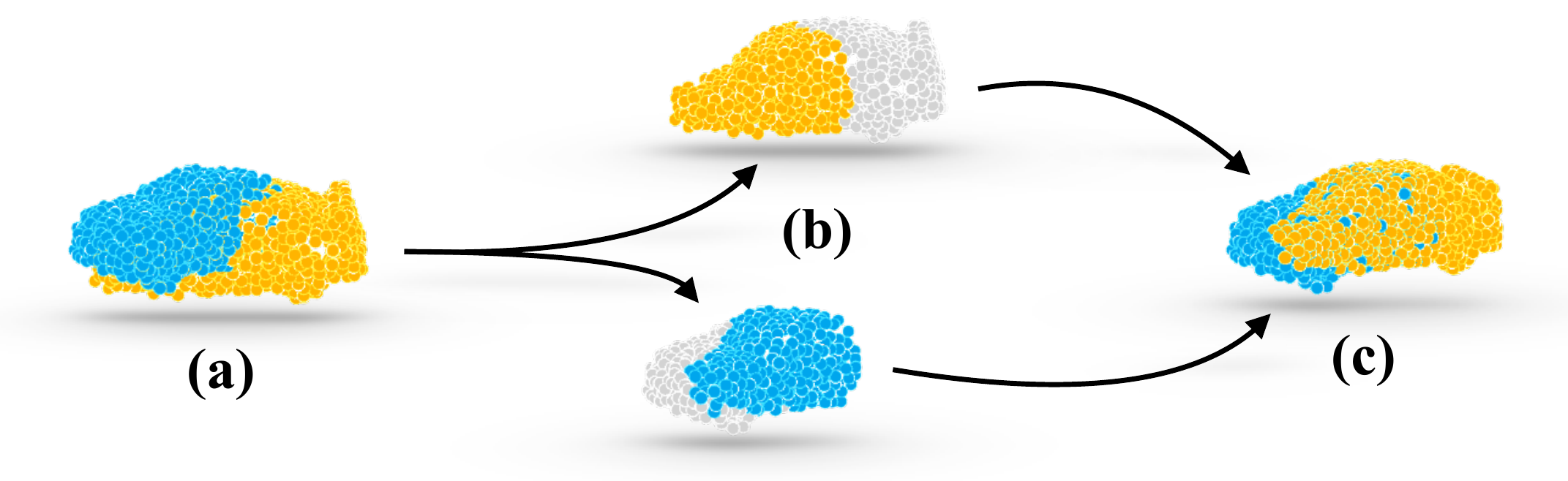}
    \vspace{-.5cm}
    \caption{Given (a) input partial point clouds, \ourmethod detects (b) the overlap regions that are then used for the estimation of (c) the transformation that aligns the two point clouds.
    The non-overlap regions in (b) are shown in grey. 
    Our approach focuses on the geometric information in the overlap regions to perform the point cloud registration.}
    \label{fig:overlap_example}
\end{figure}

Fig.~\ref{fig:overlap_example} shows an example of registration where the overlap regions that our approach can automatically determine are colored with the respective point cloud colors (non-overlap regions are in grey).
This example depicts a case where the overlap between the two point clouds is 50\%.

\subsection{Training}\label{sec:training}
We jointly optimize three tasks, finding overlap regions, clustering point clouds, and estimating rigid transformations.

\vspace{.1cm}
\noindent\textbf{Overlap score loss.} 
The goal of the overlap score loss is to detect the overlap region between $\bm{\mathcal{P}}$ and $\bm{\mathcal{Q}}$. Given ground-truth $\bar{\bm{T}}$,
the ground-truth $\bar{\bm{o}}_{p_i}$ of $\bm{p}_i$ is defined
\vspace{-.2cm}
\begin{equation}\label{eq:gtop}
\vspace{-.2cm}
	\bar{\bm{o}}_{p_i} = 
	\begin{cases}
		1, & \left(\min_{q_j\in \bm{\mathcal{Q}}}\|\bar{\bm{T}}(\bm{p}_i) -\bm{q}_j\|\right) < \eta \\
		0, & \text{otherwise}
	\end{cases},
\end{equation}
$\bar{\bm{o}}_{q_j}$ is calculated in the same way. 
The overlap score loss is defined as $\mathcal{L}_{O} = \frac{1}{2}\left(\mathcal{L}_{\bm{\mathcal{P}}} + \mathcal{L}_{\bm{\mathcal{Q}}} \right)$, where
\vspace{-0.2cm}
\begin{equation*}
\vspace{-.2cm}
	\mathcal{L}_{\bm{\mathcal{P}}} = -\frac{1}{|\bm{\mathcal{P}}|}\sum_{i}\left(\bar{\bm{o}}_{\bm{p}_i}\log \bm{o}_{\bm{p}_i}+\left(1-\bar{\bm{o}}_{\bm{p}_i}\right)\log\left(1-\bm{o}_{\bm{p}_i}\right)\right).
\end{equation*}
By the same operation,  we can attain $\mathcal{L}_{\bm{\mathcal{Q}}}$. 

\vspace{.1cm}
\noindent\textbf{Global registration loss.}\label{subs:clus}
We first exploit the registration error-based loss function to train our model. To handle the partially-overlapping problem well, we adopt a robust error metric that does not incur high computational overhead. Specifically, we formulate the registration loss as
\begin{equation*}
    \mathcal{L}_g =\sum_{\hat{\bm{p}}\in\bm{\mathcal{\hat{P}}}}\psi_{\nu}\left(\mathcal{D}\left(\hat{\bm{p}},m\left(\bar{\bm{T}}(\bm{p}),\bm{\mathcal{Q}}\right)\right)\right).
    \vspace{-.2cm}
\end{equation*}
Here, $m(\bm{x}, \bm{\mathcal{Q}})$ maps point $\bm{x}$ to its nearest
point in $\bm{\mathcal{Q}}$. $\bm{\mathcal{\hat{P}}}$ denotes the transformed $\bm{\mathcal{P}}$ using the estimated transformation. $\mathcal{D}\left(\cdot,\cdot\right)$ defines as the Euclidean distance of two vectors. $\psi_{\nu}$ is the Welsch’s function~\cite{holland1977robust} as $\psi_{\nu}\left(x\right)=1-\exp\left(-\frac{x^2}{2\nu^2}\right)$. $\nu > 0$ is a user-specified parameter.

\vspace{.1cm}
\noindent\textbf{Clustering-based loss.}
To ensure that our network learns a consistent GMM representation across feature and geometric space rather than fit a GMM in a single feature space. We introduce the following loss.
\vspace{-.2cm}
\begin{equation}
\begin{aligned}
    \mathcal{L}_{c} = &-\sum_{ij}\gamma^p_{ij}\log \frac{\exp{\left(-\mathcal{D}\left(\bm{p}_i,\bm{\mu^p_j}\right)\right)}}{\sum_l\exp{\left(-\mathcal{D}\left(\bm{p}_i,\bm{\mu^p_l}\right)\right)}} \\
    & -\sum_{ij}\gamma^q_{ij}\log \frac{\exp{\left(-\mathcal{D}\left(\bm{q}_i,\bm{\mu^q_j}\right)\right)}}{\sum_l\exp{\left(-\mathcal{D}\left(\bm{q}_i,\bm{\mu^q_l}\right)\right)}}.
\end{aligned}
\end{equation}


\section{Experiments}
We perform extensive experiments and ablation studies on both synthetic and real-world point cloud datasets, including the ModelNet40 \cite{wu20153d}, 7Scenes \cite{shotton2013scene} and ICL-NUIM \cite{handa2014benchmark}. 
Unless otherwise specified, $J=72,K=5,\eta=0.1$.
Our implementation is built on the PyTorch library. 
We used AdamW as optimizer with a base learning rate of 0.001. 
The batch size is 32, and the learning rate was reduced by a factor of 0.7 every 20 epochs. 
We trained our model for 200 epochs.
All of our models were trained on two Tesla V100-PCI-E-32G GPUs.
We set GMM components $L=48$. 
To alleviate the local minima clustering solution, we initialize the centroids of Wasserstein K-Means using the farthest point sampling strategy and the equal partition constraint.

\subsection{Comparisons}
We compare our approach with methods from state-of-the-art by running their code on the selected datasets.
Because \ourmethod belongs to the category of learning-based probabilistic registration, we compare its performance with probabilistic approaches: 
CPD~\cite{myronenko2010point}, 
GMMReg~\cite{jian2010robust}, 
SVR~\cite{campbell2015adaptive}, 
DeepGMR~\cite{yuan2020deepgmr}, 
and FilterReg~\cite{gao2019filterreg}. 
For these traditional methods, we use the implementations provided by probreg~\cite{kenta2019probreg}. We improve the DeepGMR by replacing its encoder of DeepGMR with our encoder for a fair comparison. 
We also include point-level correspondence methods: traditional methods 
(i.e., ICP \cite{besl1992method}, FGR~\cite{zhou2016fast}), using implementations from Open3D~\cite{zhou2018open3d}, 
and the learning-based state-of-the-art methods 
RGM~\cite{fu2021robust}, OMNET~\cite{xu2021omnet}, RIENet~\cite{shen2022reliable}, and REGTR~\cite{yew2022regtr} (using authors' implementation).

\subsection{Evaluation metrics} 
We evaluate the registration quality by using the \textit{Mean Absolute Error (MAE)} between the estimated rotation angle $\theta_{est}$ and ground truth $\theta_{gt}$, and the estimated translation $t_{est}$ and ground truth $t_{gt}$ \cite{fu2021robust,wang2019deep}.
Rotation metrics are in \textit{degrees}, while translation metrics are in \textit{cm}.
We also use \textit{Clip Chamfer Distance (CCD)}, which measures how close the two point clouds are aligned with each other \cite{fu2021robust}.
To avoid the influence of outliers in partial-to-partial registration, each pair of points whose distance is larger than 0.1 is discarded.
This is implemented by setting the threshold $d = 0.1$. 




\subsection{Datasets}\label{subsec:data}

\noindent\textbf{ModelNet40}~\cite{wu20153d} contains 12,311 meshed CAD models from 40 categories. 
Following~\cite{huang2022unsupervised}, we split the dataset and create two setups, namely \textit{same-category} and \textit{cross-category}.
The same-category setup contains 20 categories for training and testing. 
The cross-category setups consist of 20 categories that are disjoint between training and testing for the generalization performance evaluation. 
Following RGM \cite{fu2021robust}, each category consists of official train/test splits. 
To select models for evaluation, we take 80\%, and 20\% of the official train split as the training set and validation set, respectively, and the official test split for testing. In realistic scenes, the points in $\bm{\mathcal{P}}$ have no exact correspondences in $\bm{\mathcal{Q}}$ \cite{xu2021omnet}. We thus uniformly sample 1,024 points from each CAD model two times with different random seeds to generate $\bm{\mathcal{P}}$ and $\bm{\mathcal{Q}}$, breaking exact correspondences between the input point clouds, which is different from previous works. Following previous works \cite{huang2020feature,fu2021robust}, we randomly draw a rigid transformation along each axis to generate the target points; the rotation along each axis is sampled in $ [0, 45^\circ] $ and translation is in $[-0.5, 0.5]$. 
$\nu=0.1$.

\noindent\textbf{7Scenes}~\cite{shotton2013scene} is a 3D dataset captured in seven indoor scenes by using a Kinect RGB-D camera and consists of seven scenes: Chess, Fires, Heads, Office, Pumpkin, Red kitchen, and Stairs.
7Scenes is split into two parts, one is for training with 296 scans and one is for testing with 57 scans. 
7Scenes is widely used to assess registration performance with real-world data. 
The point clouds are also uniformly sampled two-time from the original point clouds and a rigid transformation for one of the two samples to simulate the pose difference. 
We randomly draw a rigid transformation along each axis to generate the target points; the rotation along each axis is sampled within the interval $[0, 45^\circ]$, while the translation is sampled within the interval $[-0.5, 0.5]$. 
$\nu=0.5$. 

\noindent\textbf{ICL-NUIM}~\cite{handa2014benchmark} is derived from RGB-D scans in the Augmented ICL-NUIM dataset~\cite{choi2015robust}. Following~\cite{yuan2020deepgmr}, we split the scenes in ICL-NUIM into 1,278 and 200 scans for training and testing, respectively.
$\nu=0.5$. 

\begin{table}[t]
\centering
\caption{Partial-to-Partial Registration results on ModelNet40.}
\label{tb:mncn}%
\resizebox{1\linewidth}{!}{%
\begin{tabular}{l|c c c|c c c}
\toprule
\multirow{2}{*}{Method} &
\multicolumn{3}{c|}{Same-category setup} &
\multicolumn{3}{c}{Cross-category setup} \\
& MAE(R) & MAE($\bm{t}$) & CCD 
& MAE(R) & MAE($\bm{t}$) & CCD \\
\midrule
ICP \cite{besl1992method}
& 10.333 & 0.1034 & 0.1066 
& 11.499 & 0.1084 & 0.1142\\
FGR \cite{zhou2016fast}
& 22.103 & 0.1273 & 0.1108 
& 21.928 & 0.1281 & 0.1175\\
RGM \cite{fu2021robust}
& 0.8211 & 0.0094 & 0.0729
& 1.3249 & 0.0164 & 0.0832\\
OMNET~\cite{xu2021omnet} 
& 2.5944 & 0.0249 & 0.0936
& 3.6001 & 0.0355 & 0.1063 \\
RIENet \cite{shen2022reliable}
& 4.9586 & 0.0152 & 0.0735
& 5.5074 & 0.0425 & 0.1072\\
REGTR~\cite{yew2022regtr}
& 0.7836 & \bf0.0066 & 0.0676
& 0.9105 & 0.0071 & 0.0645\\
\hline
CPD~\cite{myronenko2010point} 
& 11.033 & 0.1139 & 0.1110
& 12.681 & 0.1153 & 0.1154\\
GMMReg~\cite{jian2010robust} 
& 13.677 & 0.1344 & 0.1180
& 14.899 & 0.1440 & 0.1268\\
SVR \cite{campbell2015adaptive}  
& 11.857 & 0.1162 & 0.1170 
& 13.120 & 0.1225 & 0.1237\\
FilterReg \cite{gao2019filterreg} 
& 20.363 & 0.1558 & 0.1182
& 20.531 & 0.1646 & 0.1302\\
DeepGMR \cite{yuan2020deepgmr}
& 6.8043 & 0.0683 & 0.1182
& 7.3139 & 0.0718 & 0.1207\\
\ourmethod (ours)
& \bf0.5892 & 0.0079 & \bf0.0493  
& \bf0.6309 & \bf0.0055 & \bf0.0548\\
\bottomrule
\end{tabular}
}
\end{table}

\subsection{Evaluation on ModelNet40}\label{sec:dm}

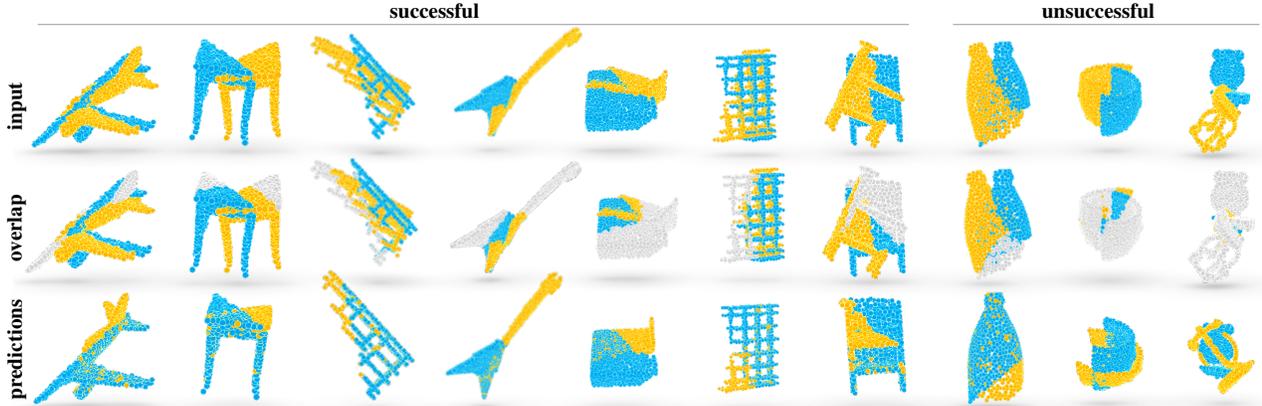
\begin{figure*}[t]
	\centering
    \begin{overpic}[width=2\columnwidth]{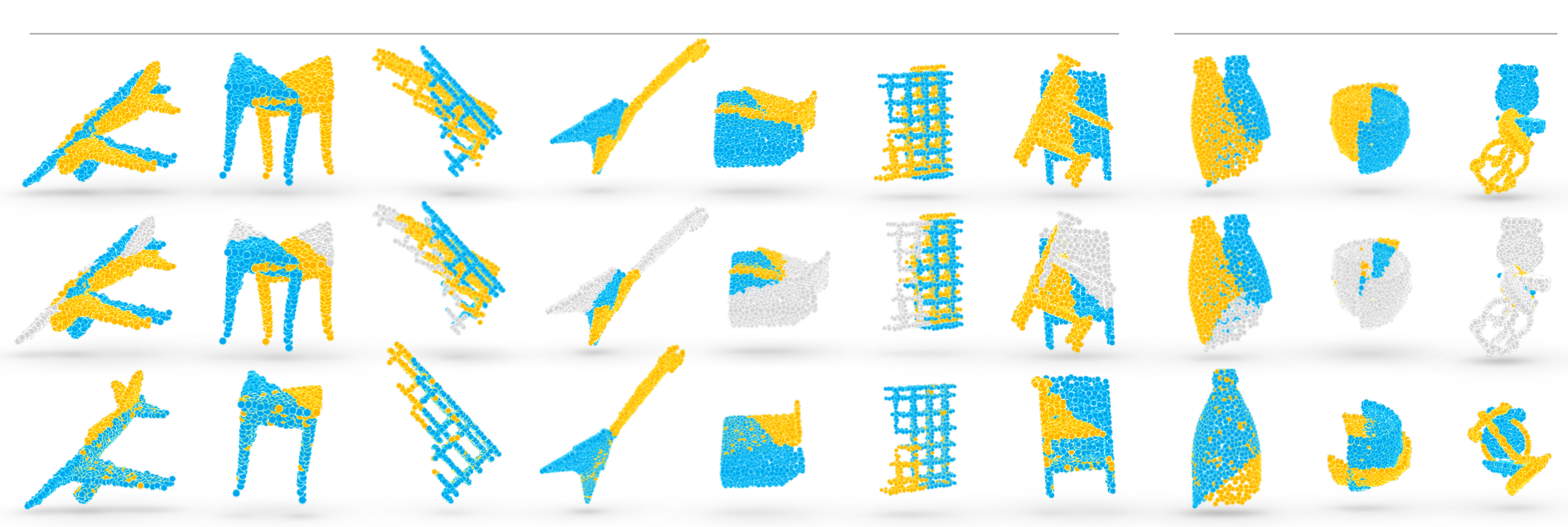}
    \put(-0.5,23){\color{black}\footnotesize\rotatebox{90}{\textbf{input}}}
    \put(-0.5,12.5){\color{black}\footnotesize\rotatebox{90}{\textbf{overlap}}}
    \put(-0.5,1.5){\color{black}\footnotesize\rotatebox{90}{\textbf{predictions}}}
    \put(30, 32){\color{black}\footnotesize\textbf{successful}}
    \put(82, 32){\color{black}\footnotesize\textbf{unsuccessful}}
    \end{overpic}
	\caption{Successful and unsuccessful registration results on ModelNet40 using \ourmethod. The non-overlap regions are shown in grey.}
	\label{fig:mndvs}
\end{figure*}

\noindent\textbf{Same-category setup.}
We follow the protocol in \cite{yew2022regtr} to generate partial-to-partial point cloud pairs, which are closer to real-world applications. 
We first generate a half-space with a random direction for each point cloud and shift it to retain approximately 70\% of the points, i.e., 717 points. 
Tab.~\ref{tb:mncn} reports the results.
\ourmethod significantly outperforms both traditional and deep learning-based methods. 
Unlike DeepGMR, \ourmethod can better detect overlapping regions thanks to the newly introduced overlap scores.
Fig.~\ref{fig:mndvs} shows various successful and unsuccessful registration results.
We can observe that cases with low overlap (e.g., guitar) and with repetitive structures (e.g., net) can be handled by \ourmethod.
The most frequent unsuccessful cases involve point clouds of symmetric objects that can have multiple correct registration solutions along the symmetry axis that do not match with the ground-truth one.
This is an intrinsic problem of symmetric cases \cite{xu2021omnet}.
The second most frequent unsuccessful cases involve point cloud pairs with repetitive local geometric structures.
These make GMM clustering underperform because features of similar structures in different locations have a small distance in the feature space.


\begin{table}[t]
\centering
\caption{Registration results on ModelNet40 with jittering noise or density variation.
}
\label{tb:noise}%
\resizebox{1\linewidth}{!}{%
\begin{tabular}{l|c c c|c c c}
\toprule
\multirow{2}{*}{Method} &
\multicolumn{3}{c|}{Jittering} &
\multicolumn{3}{c}{Density variation} \\
& MAE(R) & MAE($\bm{t}$) & CCD 
& MAE(R) & MAE($\bm{t}$) & CCD \\
\midrule
ICP \cite{besl1992method}
& 10.609 & 0.1058 & 0.1087 
& 5.6654 & 0.0638 & 0.0908\\
FGR \cite{zhou2016fast}
& 24.534 & 0.1413 & 0.1168
& 16.867 & 0.0295 & 0.0890\\
RGM \cite{fu2021robust}
& 1.3536 & 0.0261 & 0.0447
& 1.5948 & 0.0103 & 0.0672\\
OMNET~\cite{xu2021omnet} 
& 2.6996 & 0.0259 & 0.0977 
& 3.1372 & 0.0313 & 0.1010 \\
RIENet \cite{shen2022reliable}
& 5.8212 & 0.0635 & 0.1003
& 5.5928 & 0.0725 & 0.1059\\
REGTR~\cite{yew2022regtr}
& 1.0984 & 0.0080 & 0.0732
& 3.6409 & 0.0237 & 0.0946\\
\hline
CPD~\cite{myronenko2010point} 
& 11.049 & 0.1141 & 0.1120
& 8.5349 & 0.0793 & 0.0941\\
GMMReg~\cite{jian2010robust} 
& 13.763 & 0.1343 & 0.1187
& 8.1247 & 0.0853 & 0.1024\\
SVR \cite{campbell2015adaptive}  
& 12.174 & 0.1192 & 0.1178
& 5.0848 & 0.0594 & 0.0949\\
FilterReg \cite{gao2019filterreg} 
& 19.921 & 0.1548 & 0.1184
& 20.098 & 0.1016 & 0.1109\\
DeepGMR \cite{yuan2020deepgmr} 
& 8.8242 & 0.0699 & 0.1210
& 7.1273 & 0.0689 & 0.1206\\
\ourmethod (ours) 
& \bf0.9111 & \bf0.0071 & \bf0.0645 
& \bf1.3523 & \bf0.0100 & \bf0.0534 \\
\bottomrule
\end{tabular}
}
\end{table}

\vspace{.1cm}
\noindent\textbf{Cross-category setup.}
To evaluate the generalization ability of our method, we train our model and all the other deep learning-based methods on 20 categories (i.e., airplane, bathtub, bed, bench, bookshelf, bottle, bowl, car, chair, cone, cup, curtain, desk, door, dresser, flower pot, glass box, guitar, keyboard, and lamp), and test them on other different 20 categories (i.e., laptop, mantel, monitor, nightstand, person, piano, plant, radio, range hood, sink, sofa, stairs, stool, table, tent, toilet, TV stand, vase, wardrobe, and Xbox). 
The data pre-processing is the same as that of the first experiment.
Tab.~\ref{tb:mncn} shows that the performances of all learning-based methods are worse than those trained in the same categories.
\ourmethod still outperforms the other baselines in this setup.

\noindent\textbf{Noisy shapes.} 
To evaluate the robustness of \ourmethod, Gaussian noise sampled from $\mathcal{N}(0, 0.01)$ and clipped to $[-0.05, 0.05]$ is independently added to each point coordinate.
We train and test our \ourmethod on the noisy data of ModelNet40.
From the results listed in the left part of Tab.~\ref{tb:noise} (Jittering), we can observe that \ourmethod can effectively handle noise.
This occurs thanks to our cluster-based network that can extract more robust features.

\noindent\textbf{Density variation.}
We aim to evaluate the robustness of methods in handling variations of point cloud densities.
We randomly discard points of one point cloud and repeat the remaining points to keep the same number of points to generate a pair of point clouds with density differences. 
Tab.~\ref{tb:noise} (Density variation) reports the results when the point density is reduced to 50\%.
\ourmethod can outperform the other methods when source and target point clouds have different densities. 
This occurs mainly because the source and target point clouds are modeled as GMMs, therefore the alignment between the two GMMs becomes less sensitive to density, as opposed to establishing the explicit point correspondence.

\subsection{Evaluation on 7Scenes and ICL-NUIM} 
We conduct experiments on two indoor scenes: real-world 7Scenes \cite{shotton2013scene} and ICL-NUIM.
The point clouds are uniformly sampled from the original point clouds to generate source and target counterparts.
Following~\cite{huang2022unsupervised}, we resample 2,048 points from each model two times with different random seeds to generate $\bm{\mathcal{P}}$ and $\bm{\mathcal{Q}}$, then generate a half-space with a random direction for each point cloud and shift it to retain approximately 70\% of the points (1,433) to generate the partial data. As shown in Tab.~\ref{tb:indoor}, \ourmethod achieves the lowest errors on all the metrics on both datasets. 
\ourmethod outperforms DeepGMR and GMMReg thanks to the clustered attention network to detect overlapping regions and produce more distinctive features.

\begin{table}[t]
\vspace{-.2cm}
\small
\centering
\caption{The registration results on 7Scenes and ICL-NUIM.
	} 
	\resizebox{1\linewidth}{!}{%
	\begin{tabular}{l|c c c|c c c}
		\toprule
		\multirow{2}{*}{Method} &
		\multicolumn{3}{c|}{7Scenes} &
		\multicolumn{3}{c}{ICL-NUIM} \\
        & MAE(R) & MAE($\bm{t}$) & CCD 
        & MAE(R) & MAE($\bm{t}$) & CCD \\
		\midrule
        ICP \cite{besl1992method}
        & 18.266 & 0.2346 & 0.0793
        & 10.539 & 0.3301 & 0.1410\\
        FGR \cite{zhou2016fast}
        & 1.1736 & 0.0198 & 0.0270
        & 0.8792 & 0.0332 & 0.0835\\
        \hline
        RGM \cite{fu2021robust}
        & 3.0334 & 0.0445 & 0.0425
        & 1.3279 & 0.0416 & 0.0840\\
        OMNET~\cite{xu2021omnet} 
        & 9.8499 & 0.1416 & 0.1879
        & 17.177 & 0.4587 & 0.1947 \\
        REGTR~\cite{yew2022regtr}
        & 4.6143 & 0.0827 & 0.1733
        & 3.2503 & 0.1044 & 0.0902\\
        \hline
        CPD~\cite{myronenko2010point} 
        & 4.9897 & 0.1056 & 0.0900
        & 9.8322 & 0.3828 & 0.1581\\
        GMMReg~\cite{jian2010robust} 
        & 11.081 & 0.1213 & 0.1374
        & 6.5411 & 0.1700 & 0.1530\\
        SVR \cite{campbell2015adaptive}  
        & 10.729 & 0.1152 & 0.1388
        & 6.3229 & 0.1946 & 0.1528\\
        FilterReg \cite{gao2019filterreg} 
        & 18.113 & 0.2521 & 0.0636
        & 28.317 & 0.7930 & 0.1693\\
        DeepGMR \cite{yuan2020deepgmr} 
        & 8.8478 & 0.1534 & 0.0541
        & 6.4600 & 0.1899 & 0.1269\\
        \ourmethod  (ours)
        & \bf0.5764 & \bf0.0088 & \bf0.0214
        & \bf0.6279 & \bf0.0305 & \bf0.0732\\
		\bottomrule
	\end{tabular}
 }
	\label{tb:indoor}
\end{table}

\vspace{-.1cm}
\subsection{Ablation Studies}
\vspace{-.1cm}
\noindent\textbf{Components of \ourmethod.}
We analyze the effectiveness of \ourmethod components in the case of partial-to-partial registration (same-category setup).
We assess the three key novel components of \ourmethod: Spherical Positional Encoding (SPE), Cluster-based Self-Attention (CSA), and Overlap Score Prediction (OSP). 
Tab.~\ref{tb:component} shows that \ourmethod underperforms when the overlap scores are not used.
CSA and SPE modules help achieving a higher registration accuracy, as they allow for more distinctive features to be produced.

\begin{table}[t]
\vspace{-.2cm}
\small
\centering
\caption{Ablation study on ModelNet40.}
\resizebox{0.65\columnwidth}{!}{
\begin{tabular}{cccccc}
\toprule
SPE & CSA & OSP & MAE(R) & MAE($\bm{t}$) & CCD\\
\midrule
& \checkmark & \checkmark  & 0.9534 & 0.0096 & 0.0589\\
\checkmark &  & \checkmark & 1.9060 & 0.0169 & 0.0889\\
\checkmark & \checkmark &  & 6.7087 & 0.0729 & 0.1155\\
\checkmark & \checkmark & \checkmark & \bf0.5892 & \bf0.0079 & \bf0.0493\\
\bottomrule
\end{tabular}
}
\label{tb:component}
\end{table}

\vspace{.1cm}
\noindent\textbf{Loss functions.} 
We train our model with different combinations of the Global Registration loss (GR), the Clustering-based loss (GS), and the Overlap Score loss (OS).
Experiments are conducted on ModelNet40 (same-category setup). 
Tab.~\ref{tb:loss} shows that the combination of GR and OS losses provides the major contribution. 

For the ablation study results of overlapping ratios and clusters, please refer to the supplementary material.

\begin{table}[t]
\vspace{-.2cm}
\small
\centering
\caption{Loss function analysis on ModelNet40.}
\label{tb:loss}
\resizebox{0.65\linewidth}{!}{
\begin{tabular}{c c c | c c c c}
\toprule
GR & CL & OS & MAE(R) & MAE($\bm{t}$) & CCD\\
\midrule
\checkmark   & ~ & ~ & 3.7284 & 0.0272 & 0.0958\\
& \checkmark &   & 7.0198 & 0.0553 & 0.1133\\
&            & \checkmark & 4.4373 & 0.0392 & 0.1006\\
\checkmark & \checkmark &  & 3.2335 & 0.0262 & 0.0939\\
& \checkmark & \checkmark & 2.6470 & 0.0250 & 0.0751\\
\checkmark & & \checkmark & 0.7828 & 0.0085 & 0.0515\\
\checkmark & \checkmark & \checkmark & \bf0.5892 & \bf0.0079 & \bf0.0493\\
\bottomrule
\end{tabular}
}
\end{table}

\vspace{.1cm}
\noindent\textbf{Inference time.}
We evaluate the efficiency of \ourmethod and compare it to other approaches on ModelNet40 (same-category setup).
We average the inference time of \ourmethod using a single Tesla V100 GPU (32G) and two Intel(R) 6226 CPUs. \textit{f} and \textit{c} represent the full and cluster-based attention.
Tab.~\ref{tb:time} reports the results. 
Compared to RGM, \ourmethod utilizes similar network architecture but different matching strategies and attention modules. \ourmethod (c) outperforms RGM and reduces about $9\times$ the computation time. Compared with full attention, our cluster-based attention can speed up $8\times$ times, which verifies the effectiveness of our clustered strategy in reducing the computing complexity.
Our method is inferior to DeepGRM because the overlap detection module is introduced to handle partial overlap cases.

\begin{table}[t]
\vspace{-.2cm}
\centering
\caption{Comparisons of the average inference time.}
\resizebox{1\linewidth}{!}{
\begin{tabular}{l|cc|cccc}
\toprule
Method  & CPD   & GMMReg & RGM & DeepGMR & \ourmethod (f) & \ourmethod (c) \\
\hline
Time(s) & 4.347 & 2.536  & 0.057  & \bf0.002  & 0.047 & 0.006\\	
\bottomrule
\end{tabular}
}
\label{tb:time}
\end{table}

\section{Conclusions}
We presented a learning-based probabilistic registration approach for partially overlapping point clouds, named \ourmethod.
We use a clustered attention-based network to detect overlap regions between two point clouds and formulate the registration problem as the minimization of the discrepancy between Gaussian mixtures under the guidance of the overlap information.
Experiments show that \ourmethod outperforms prior traditional and deep learning-based registration approaches across a variety of data setups.
\ourmethod is robust to noise and can effectively generalize to new object categories and on real-world data. 
\ourmethod features a novel way to integrate 3D neural networks inside a probabilistic registration paradigm. 
Future research directions include the development of an unsupervised method to detect the overlap regions to reduce the dependence on labeled data.

{\small
\bibliographystyle{ieee_fullname}
\bibliography{egbib}
}

\newpage
~~~~~~~

\newpage
\appendix
\section{Supplementary Material}
Firstly, we provide additional ablation results on ModelNet40 (we use the official train/test split for training and testing).
Then, we provide the results of complete-to-complete and complete-to-partial point cloud registration. 
Lastly, we show qualitative results on 7Scenes.

\subsection{Additional ablation study results}
\paragraph{Different overlapping ratios.}
Because the overlap ratio may affect the performance of registration, we analyze the performance variation when the overlap ratio decreases gradually. 
We evaluate the performance of \ourmethod on noisy ModelNet40. 
We utilize the same crop setting as RPMNet to generate point clouds with approximate overlap ratios of 70\%, 60\%, 50\%, 40\%, and 30\%, respectively. 
Based on \cite{huang2020feature,fu2021robust}, we randomly draw a rigid transformation along each axis to generate the target points; the rotation along each axis is sampled in $ [0, 45^\circ] $ and translation is in $[-0.5, 0.5]$. We train \ourmethod on data with a 50\% overlap ratio and test on different overlap ratios. Tab.~\ref{tb:ratio} shows the registration results under different overlap ratios with Gaussian noise sampled from $\mathcal{N}(0,0.01)$ and clipped to $[-0.05, 0.05]$. 
The lower the overlap ratio, the higher the registration error.

\begin{table}[b]
\vspace{-.3cm}
\centering
\caption{The effects of the overlap ratio on ModelNet40.}
\resizebox{0.8\linewidth}{!}{
\begin{tabular}{l|c c c c c c}
\toprule
Ratio           & 30\%   & 40\%    & 50\%    & 60\%    & 70\%\\
\hline
MAE (R)         & 5.6462 & 3.5995  & 2.1000  & 1.2067  & 0.9111\\	
MAE ($\bm{t}$)  & 0.1220 & 0.0716  & 0.0178  & 0.0159  & 0.0071\\	
CCD             & 0.1097 & 0.0957  & 0.0937  & 0.0664  & 0.0645\\	
\bottomrule
\end{tabular}
}
\label{tb:ratio}
\end{table}

\paragraph{Different cluster numbers.}
We show the performances of \ourmethod under different cluster numbers (8, 16, 32, 48, and 64) with a 50\% overlap ratio.
Tab.~\ref{tb:cluster} shows that when the number of clusters is small (e.g.~8, 16) we obtain a larger registration error.
When we set the number of clusters between 32 to 64, the registration error is reduced and it varies slightly.

\begin{table}[b]
\small
\vspace{-.3cm}
\centering
\caption{The effects of the cluster numbers on ModelNet40 with 50\% overlapping ratio and Gaussian noise.}
\resizebox{0.8\linewidth}{!}{
\begin{tabular}{l|c c c c c}
\toprule
Ratio           & 8      & 16      & 32      & 48      & 64\\
\hline
MAE (R)         & 5.6462 & 3.5995  & 2.1625 & 2.0834 & 2.1000\\	
MAE ($\bm{t}$)  & 0.1220 & 0.0716  & 0.0187 & 0.0181 & 0.0178\\	
CCD             & 0.1097 & 0.0957  & 0.0942 & 0.0885 & 0.0937\\	
\bottomrule
\end{tabular}
}
\label{tb:cluster}
\end{table}

\subsection{Additional registration results}
Because \ourmethod is a GMM-based method, we compare it against GMM-based baselines.
\vspace{.1cm}

\paragraph{Complete-to-complete setup.} 
We first evaluate the complete-to-complete registration performance on ModelNet40 with Gaussian noise sampled from $\mathcal{N}(0,0.01)$ and clipped to $[-0.05, 0.05]$ and follow the sampling and transformation settings in Sec.~\ref{subsec:data}. 
Tab.~\ref{tb:mncc} shows that our method can outperform the GMM-based baselines on this setup.
\begin{table}[h]
\vspace{-.3cm}
\centering
\caption{Registration results on ModelNet40.}
\label{tb:mncc}%
\resizebox{1\linewidth}{!}{%
\begin{tabular}{l|c c c|c c c}
\toprule
\multirow{2}{*}{Method} &
\multicolumn{3}{c|}{Complete-to-complete setup} &
\multicolumn{3}{c}{Complete-to-partial setup} \\
& MAE(R) & MAE($\bm{t}$) & CCD 
& MAE(R) & MAE($\bm{t}$) & CCD \\
\midrule
CPD~\cite{myronenko2010point} 
& 0.8171 & 0.0050 & 0.0037
& 10.293 & 0.0767 & 0.1118\\
GMMReg~\cite{jian2010robust} 
& 7.7326 & 0.0508 & 0.0837
& 24.318 & 0.2578 & 0.1119\\
SVR \cite{campbell2015adaptive}  
& 7.8047 & 0.0592 & 0.0744 
& 24.063 & 0.2480 & 0.0947\\
FilterReg \cite{gao2019filterreg} 
& 3.4899 & 0.0247 & 0.0605
& 30.653 & 0.2676 & 0.1197\\
DeepGMR \cite{yuan2020deepgmr}
& 2.2736 & 0.0150 & 0.0503
& 12.612 & 0.1527 & 0.1266\\
\ourmethod (ours)
& \bf0.1461 &\bf 0.0021 & \bf 0.4237  
& \bf7.2820 &\bf 0.0633 & \bf 0.1142\\
\bottomrule
\end{tabular}
}
\end{table}

\vspace{.1cm}
\paragraph{Complete-to-partial setup.} 
We also evaluate the complete-to-partial registration performance on ModelNet40 with Gaussian noise. We crop the generated source point cloud in Sec.~\ref{subsec:data} to create a new source point cloud with approximate overlap ratios of 70\%, which includes 717 points. 
We randomly draw a rigid transformation along each axis to transform the target point cloud, which contains 1024 points. 
Tab.~\ref{tb:mncc} shows that our method can outperform the GMM-based baselines also on this setup.

\begin{figure}[!hbt]
	\centering
    \begin{overpic}[width=0.99\columnwidth]{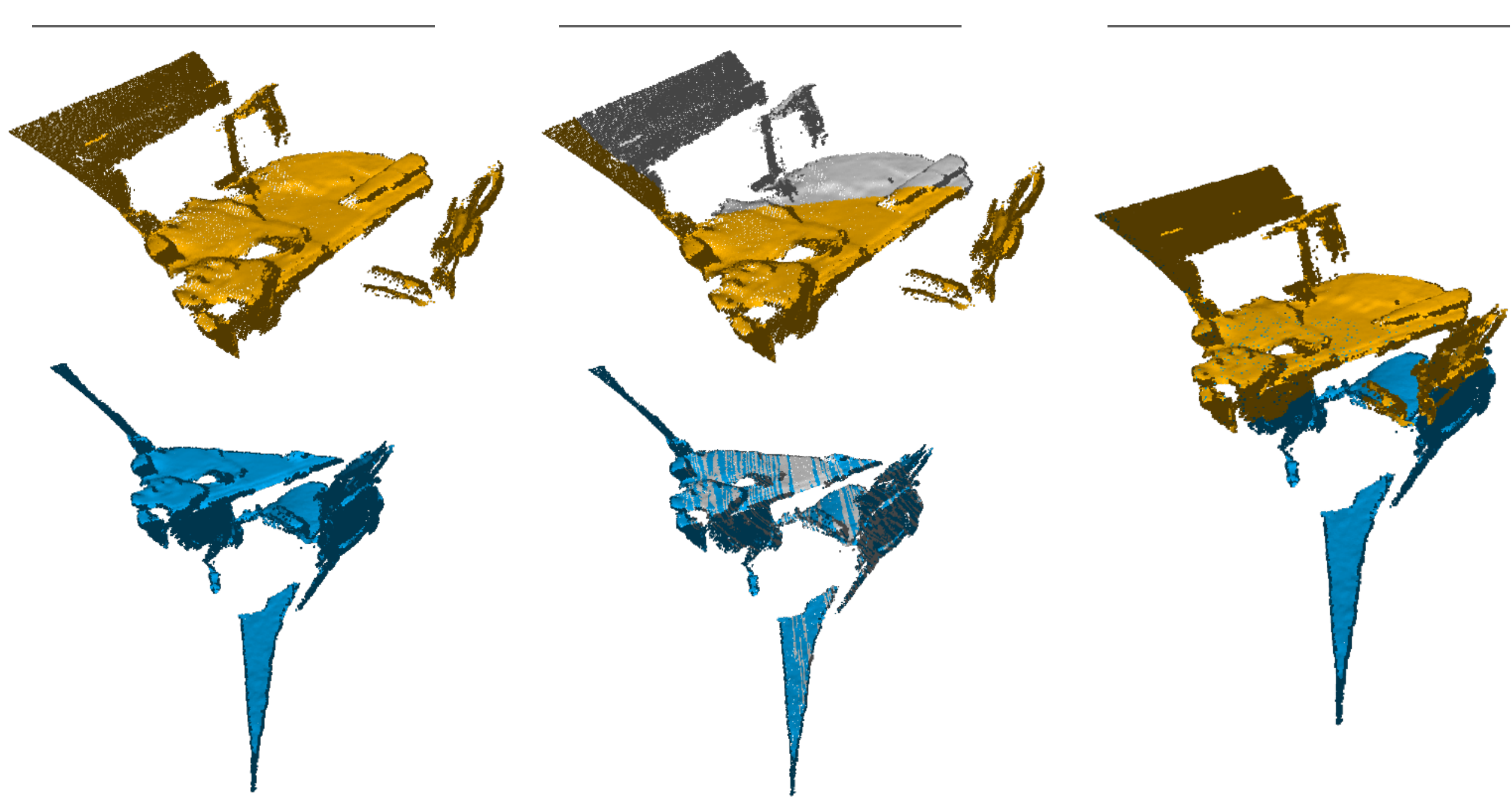}
    \put(-1.1,36){\color{black}\footnotesize\rotatebox{90}{\textbf{source}}}
    \put(-1.1,12.5){\color{black}\footnotesize\rotatebox{90}{\textbf{target}}}
    \put(100,24){\color{black}\footnotesize\rotatebox{90}{\textbf{successful}}}
    \put(12, 52){\color{black}\footnotesize{\textbf{input}}}
    \put(45, 52){\color{black}\footnotesize{\textbf{overlap}}}
    \put(82, 52){\color{black}\footnotesize{\textbf{results}}}
    \end{overpic}
    \begin{overpic}[width=0.99\columnwidth]{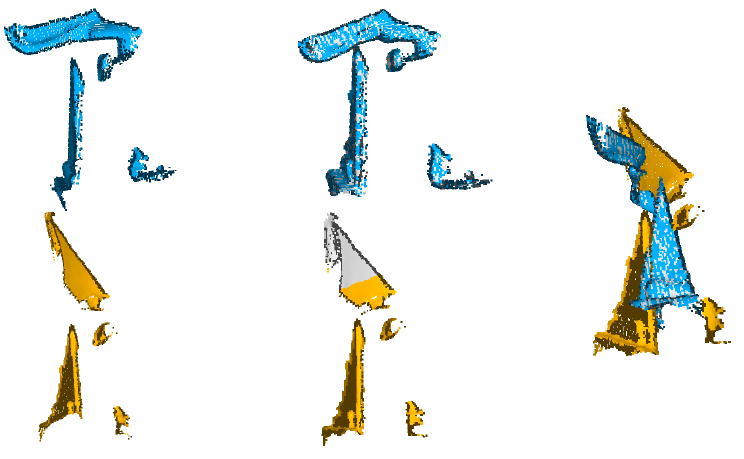}
    \put(-1.1,36){\color{black}\footnotesize\rotatebox{90}{\textbf{source}}}
    \put(-1.1,12.5){\color{black}\footnotesize\rotatebox{90}{\textbf{target}}}
    \put(100,24){\color{black}\footnotesize\rotatebox{90}{\textbf{unsuccessful}}}
    \end{overpic}
    \vspace{-.2cm}
	\caption{Qualitative results on the 7scenes dataset.}
	\label{fig:7scenes}
\end{figure}
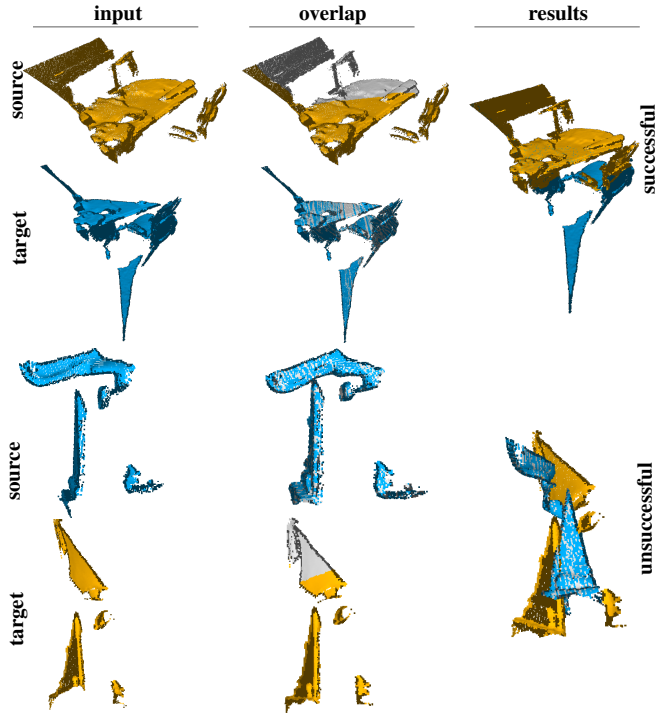

\paragraph{Visualization of predicted overlap scores and registration results on 7scene.}
Fig.~\ref{fig:7scenes} shows successful and unsuccessful registration results, where the overlap between the two point clouds is 70\%. 
For the unsuccessful case, pairs with repetitive local geometric structures lead to features of similar structures in different locations have a small distance in the feature space

\end{document}